\documentclass[conference]{IEEEtran}
\IEEEoverridecommandlockouts
\usepackage{cite}
\usepackage{amsmath,amssymb,amsfonts}
\usepackage{algorithm}
\usepackage{algpseudocode}

\usepackage{graphicx}
\usepackage{textcomp}
\usepackage{xcolor}
\usepackage{hyperref}
\def\BibTeX{{\rm B\kern-.05em{\sc i\kern-.025em b}\kern-.08em
    T\kern-.1667em\lower.7ex\hbox{E}\kern-.125emX}}
\begin{document}

\title{Graph Neural Networks in Computer Vision - Architectures, Datasets and Common Approaches}

\author{\IEEEauthorblockN{Maciej Krzywda\IEEEauthorrefmark{1}, Szymon \L ukasik\IEEEauthorrefmark{1}\IEEEauthorrefmark{3}, Amir H. Gandomi\IEEEauthorrefmark{2}}
	\IEEEauthorblockA{\IEEEauthorrefmark{1}Faculty of Physics and Applied Computer Science, AGH University of Science and Technology\\
		al.\ Mickiewicza 30, 30-059 Krak\'{o}w, Poland\\
		Email: krzywda@agh.edu.pl, slukasik@agh.edu.pl\\
		\IEEEauthorrefmark{2}Faculty of Engineering and IT, University of Technology Sydney\\
		5 Broadway, Ultimo NSW 2007, Australia\\
		Email: gandomi@uts.edu.au}
			\IEEEauthorblockA{\IEEEauthorrefmark{3}Systems Research Institute, Polish Academy of Sciences\\
		ul.\ Newelska 6, 01-447 Warsaw, Poland\\
		Email: slukasik@ibspan.waw.pl}}

\maketitle

\begin{abstract}
Graph Neural Networks (GNNs) are a family of graph networks inspired by mechanisms existing between nodes on a graph. In recent years there has been an increased interest in GNN and their derivatives, i.e., Graph Attention Networks (GAT),  Graph  Convolutional  Networks  (GCN), and Graph Recurrent  Networks (GRN). An increase in their usability in computer vision is also observed. The number of GNN applications in this field continues to expand; it includes video analysis and understanding, action and behavior recognition, computational photography, image and video synthesis from zero or few shots, and many more. This contribution aims to collect papers published about GNN-based approaches towards computer vision. They are described and summarized from three perspectives. Firstly, we investigate the architectures of Graph Neural Networks and their derivatives used in this area to provide accurate and explainable recommendations for the ensuing investigations. As for the other aspect, we also present datasets used in these works. Finally, using graph analysis, we also examine relations between GNN-based studies in computer vision and potential sources of inspiration identified outside of this field.
\end{abstract}


\section{Introduction}

The area of computer vision is a very vast area, with its application spanning from the face identification of mobile phone users to the analysis of satellite images. It also employs a plethora of different techniques, based either on statistical apparatus or unconventional nature-inspired algorithms, such as neural networks. This paper aims to consolidate and analyze current knowledge on the application of Graph-Neural Networks in computer vision. We aim to identify the most common architectures and benchmark datasets.  By studying relations between different studies, we also try to identify sources of inspiration stemming from other machine learning domains. 

The paper is organized as follows, first, in Section \label{sec:gnn}, a brief overview of Graph Neural Networks and the most common architectures is provided. Section \ref{sec:architect} covers significant contributions in this field, providing a review of modifications used in the schemes of GNNs. In the next Section \ref{sec:datasets} the most popular datasets used as a point of reference for experimental studies are listed. The relations between identified contributions to the field of computer vision are studied in Section \ref{sec:inspiration}. It also captures major concepts derived from other sources, including papers outside of the computer vision domain. The paper concludes with summarizing comments and insights gathered during our studies. 

\section{Graph Neural Networks and Their Basic Forms}

Graph Neural Networks (GNNs)\cite{4700287} can model the relationship of nodes of the graph and generate its numerical representation. It can help to express and provide a more explainable representation of such data. Graphs can represent product or customer similarities in recommendation systems, agent interactions in multiagent robotics, or transceivers in a wireless communication network. Computer vision graphs, primarily formulated as Graph Neural Networks, can be an effective tool for biometrics, face, gesture, body pose recognition on video or images. In this paper, we are using GNN as the root of graphs networks family in which other features like convolutional layers or attention mechanisms can also be found.

\subsection{Graph Convolutional Networks}

Graph Convolutional Network (GCNs)\cite{Henaff2015DeepCN} \cite{Kipf2017SemiSupervisedCW} \cite{8954227} is a kind of convolutional neural network that can work directly with graphs and their structural information. Instead of having a 2D/3D array as input in convolutional neural networks, GCNs takes a graph as an input. Similar to how convolutional neural networks extract the essential information from an image to classify the image, GCNs also pass a filter over a graph, searching for important vertices and edges that can be used to classify nodes within the graph.

\begin{figure}[htbp]
    \centering
    \includegraphics[width=\columnwidth]{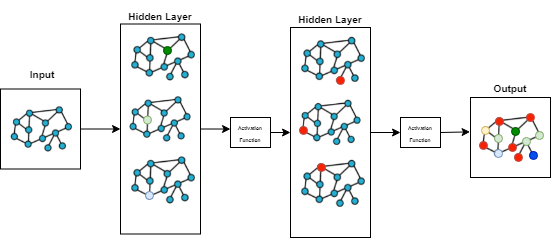}
    \caption{Graph Convolutional Network with two hidden layers.}
    \label{fig:gnn}
\end{figure}

Figure \ref{fig:gnn} demonstrates the simplified architecture of Graph Convolutional Networks that is made up of the hidden layers, and it is performing the three steps preparation, aggregation, and update. It works in sequence: input node representations are processed using a feedforward network module to produce a message, next the messages of the neighbors of each node are aggregated with respect to the edge weights, and finally, updates are combined and processed to produce the new state of the node representations.

\subsection{Graph Recurrent Networks}

Graph Recurrent Networks (GRNs), based on the simple Recurrent Neural Networks with two dense networks, achieve similar performances with the existing models for classifying the nodes on graphs. GRNs uses the random-walk transition matrix to generate the matrices sequence, feeds the output tensor into the classification and the dense attention networks, respectively, and computes the score for the labels. GRNs take a hidden state for each graph node, and it relies on an iterative message passing framework to update these hidden states in parallel. Within each iteration, neighboring nodes exchange information to absorb more global knowledge. It is different from RNNs, which require complete orders such as left-to-right orders.

\subsection{Graph Attention Networks (GATs)}

The graph attentional layer constitutes a fundamental feature of Graph Attention Networks (GATs). They were first described in work by Petar Velickovic et al. \cite{vel2018graph}. The attention mechanism is applied in a shared manner to all edges in the graph. Therefore it does not depend on upfront access to the global graph structure or features of all of its nodes (which is a limitation of many prior techniques). As opposed to GCNs, GATs assign different importance to nodes of the same neighborhood, enabling a leap in model capacity.

\section{GNN Architectures in Computer Vision}\label{sec:architect}

To analyze the main concepts present in GNNs applications for computer vision, we have conducted an extensive review of existing literature on the subject. The bibliographical query was based on the Scopus database, using citations of the original Scarcelli paper\cite{4700287} widely recognized as the source of the GNN concept. The result of the query was limited to the keywords "Computer Vision" or "Object Detection." Repetitive studies with structural advances were treated as redundant and removed from the results list, which covers 100 papers. All papers were then categorized. As the domain of computer vision lacks strict taxonomy, we have used subjective areas of the Conference on Computer Vision and Pattern Recognition (CVPR'20) conference -- one of the most prestigious venues for computer vision research. We have covered here only non-empty sub-domains and created two additional ones. "Mesh Generation and Reconstruction" and "Social Relations Detection and Recognition" were added to accommodate papers that would not fit elsewhere. This Section contains a review of the key novel architectural designs contributed within each category.

\subsection{Action and behavior recognition}
This area contains GNN application like action recognition, action segmentation\cite{Ghosh2020StackedSG},skeleton-based action recognition \cite{8954160} \cite{Huang_Huang_Ouyang_Wang_2020} \cite{DBLP:journals/corr/abs-1805-07694} \cite{DBLP:journals/corr/abs-1902-09130} \cite{DBLP:journals/corr/abs-1904-12659} \cite{Wen2019GraphCW} \cite{10.1007/978-3-030-01246-5_7} \cite{9093598} \cite{DBLP:journals/corr/abs-1911-04131}
,action classification \cite{DBLP:journals/corr/abs-1801-07455}.  Recurring architectures which we have identified will be discussed in the following subsections.

\subsubsection{Spatial-Temporal Graph Convolutional Networks (ST-GCN)}

Spatial-Temporal Graph neural Networks (STGNNs) aim to discover hidden patterns from spatial-time graphs that became important in various applications, such as recognizing human actions. The key idea of STGNNs is to take into account spatial and temporal dependence simultaneously. Current approaches integrate graph convolutions to capture spatial relationships with RNNs or CNNs for temporal modeling. In computer vision, one of the most significant progress in recent years that has drawn increasing attention is in person re-identification. In this case, a still very challenging task is to effectively overcome the problems of occlusion and the ambiguity for visually similar negative samples.
On the other hand, different frames of a video can provide complementary information for each other, and the structural information of pedestrians can provide extra discriminative cues for appearance features. So, building upon the general concept of STGNNs, J.Yang et al. \cite{Yan2020} proposed Spatial-Temporal Graph Convolutional Network (ST-GCNs), which includes two convolutional graph layers, where one is spatial and the second is temporal. The spatial part extracts structural information of a human body, while the temporal (discriminative part) mines cues from adjacent frames.

\subsubsection{Stacked Spatio-Temporal Graph Convolutional Networks}
Stacked Spatio-Temporal Graph Convolutional Networks (SPTGCN) was introduced as a solution to leverage the advantages of an encoder-decoder design for improved generalization performance and localization accuracy. Throughout the research, Ghosh et al \cite{Ghosh2020StackedSG} explored descriptors such as frame-level Visual Geometry Group(VGG), segment-level Inflated 3D ConvNets(I3D), and RCNN-based objects as node descriptors. This research aimed to enable segmentation based on joint inference over the comprehensive contextual one. SPTGCN can be applied in situations that require structured inference over long sequences with heterogeneous data types and various temporal extent.

\subsection{Biometrics, face, gesture, body pose}

Biometrics, face, gesture, body pose is one of the broadest categories which usability and approachability increases year by year. In this category we have identified research issues such as person re-identification \cite{Shen2018} \cite{Bao2019} \cite{Yan2020},  human-object interactions (HOI) \cite{DBLP:journals/corr/abs-1808-07962} \cite{Xu2019},human pose regression and estimation \cite{Zhao2019} \cite{Ci2019} \cite{Oikarinen2021} \cite{Jin2020}, face recognition \cite{Zheng2019} , human trajectory prediction \cite{Huang2019}, group emotion recognition \cite{Guo2020}, pedestrian detection\cite{Li2020}, person search \cite{DBLP:journals/corr/abs-1904-01830}, face clustering \cite{DBLP:journals/corr/abs-1904-02749}, human parsing \cite{DBLP:journals/corr/abs-1904-04536}, group activity recognition \cite{DBLP:journals/corr/abs-1904-10117}, classify perceived human emotion from gaits\cite{DBLP:journals/corr/abs-1910-12906}, crowd counting and localization\cite{DBLP:journals/corr/abs-2002-00092} and facial action unit intensity\cite{DBLP:journals/corr/abs-2004-09681}. Some architectural advances present in this area are covered below.

\subsubsection{Graph Mixture Density Network}
Graph Mixture Density Network (GraphMDNs)\cite{Oikarinen2021} are aimed at combining graph neural networks with mixture density network (MDN). An essential feature of GraphMDNs is their ability to incorporate structured graph information into a neural architecture and the possibility to model multi-modal regression targets. The main distinctive feature of GraphMDNs is the evaluation on random graphs. An additional trait is retained structural information in node representations by computing the distance between distributions of adjacent nodes.

\subsubsection{Hybrid Graph Neural Network}
Hybrid Graph Neural Network (HYGNN) were targeted to relieve the problem with crowd counting were important yet challenging task due to the large scale and density variation by interweaving the multi-scale features. In a paper prepared by Ao Luo et al. \cite{DBLP:journals/corr/abs-2002-00092} they introduce HyGNN as a solution for crowd density because graphs as a structure can improve reasoning process and knowledge representation. HYGNN integrates graphs to jointly represent nodes and two types of relations: multi-scale, capturing the feature dependencies across scales, and mutually beneficial relations building bridges for the cooperation between counting and localization. 

\subsubsection{Masked Graph Attention Networks}
Masked Graph Attention Networks (MGATs)\cite{Bao2019} operate on features extracted from a graph where nodes are able "to steer" towards other nodes' features under the guidance of label information in the form of a masks matrix. This kind of graph neural network can be used for person re-identification, where it is mainly focused on the correspondence between individual sample images and labels. MGATs can help ignore rich global mutual information residuals in the sample set.

\subsubsection{Semantic Graph Convolutional Networks}
Semantic Graph Convolutional Networks (SemGCNs)\cite{Zhao2019} were proposed as a solution for the limitation of the small receptive field of convolution filters in GCNs. GCNs in regression tasks are also characterized by a shared transformation matrix for each node in a graph. SemGCNs operate for regression tasks with graph-structured data and learn to capture semantic information to address these limitations. These semantic relations can be learned from the ground truth without additional supervision or are manually integrated into the training process. This kind of graph can be applied to 2D, and 3D human pose regression as introduced and proposed by Zhao et al. in \cite{Zhao2019}.

\subsubsection{Similarity-Guided Graph Neural Networks}
Similarity-Guided Graph Neural Networks (SGGNNs)\cite{Shen2018} creates a graph with representation (pair between samples and existing images in several galleries) and utilizes relationships to estimate visual similarities between images better. Unlike GNN approaches, SGGNN learn the edge weights directly with rich labels of gallery instance pairs, and it provides more precise information and alleviates relations' fusion. This happens during the updating process. The Probe-gallery relation feature is then constructed by the messages passing in SGGNNs, taking other nodes for similarity estimation.  

\subsubsection{Spatial-Temporal Graph Attention networks}
Spatial-Temporal Graph Attention Network (STGAT) is an architecture built on the sequence-to-sequence scheme but spatial interactions captured by graph attention network (GAT) mechanism. Operation captured spatial is running each time-step. Besides that, it adopts Long-Short Term Memory (LSTM) to encode the temporal correlations of interactions. LSTM is used, e.g., to capture the dynamic interactions of pedestrians, where the latent motions are represented with the hidden states of LSTMs, and GAT is used to aggregate hidden states of LSTMs. This solution was aimed to predict future trajectories of pedestrians and was published in \cite{DBLP:journals/corr/abs-1801-07455,DBLP:journals/corr/abs-1910-12906}.

\subsection{Medical, biological and cell microscopy}
Graph Neural Networks are also used in medical diagnosis and the medical image segmentation process. Here we have identified a single study introducing Cell Graph Convolutional Neural Networks, which usability and accuracy can boost and accelerate the detection process of cancer cell detection in medical images. 

\subsubsection{CGC-Net: Cell Graph Convolutional Network}
Cell Graph Convolutional Network (CGC-Net)\cite{9022226} are primarily used for histology image classification. A cell graph is built from an image. In GCC-Net, nuclei are regarded as the nodes and the potential cellular interactions as edges of the graph. In order to obtain a nuclear cell, CGC-Net uses a segmentation network to extract appearance features based on the segmented foreground instances. Cell Graph Convolutional Neural Networks take the whole graph as input to obtain a compact representation of the tissue micro-environment for cancer classification.

\subsection{Computational photography, image and video synthesis}
In this subfield we have identified research studies dealing with object reconstruction\cite{Fu2020}, aesthetics prediction\cite{9093412} and clustering image\cite{DBLP:journals/corr/abs-1903-11306} issues. Technical advances regarding GNNs include one architecture, covered below.

\subsubsection{RGNet}
The work \cite{Shelhamer2017} covers the RGNet architecture. It takes as input an image, representing it as a graph of local regions, and forwards it to a Fully Convolutional Network (FCN) style image aesthetics feature encoder. It is dedicated to producing task-tailored features for aesthetics assessment. The output of the FCN is a 3D feature map in which each spatial location represents one local region in the image. The spatial configuration of all spatial locations in the feature map characterizes the spatial layout of different visual elements in the image. Based on the features of local regions, a region composition graph is constructed. Each node of this graph is a local region, and any two nodes are connected by an edge-weighted by their feature similarity.

\subsection{Low-level and physics-based vision}
Following assumed categorization, we recognize low-level and physics-based vision as a separate subcategory in computer vision. In this subfield we included image restoration\cite{Qiu2020} and image generation\cite{Johnson2018ImageGF} issues. No significant advances to new architectural designs have been provided within the studied references set in this context.

\subsection{Motion and tracking}
Motion detection\cite{FENG2021116} \cite{Hu2020} and tracking \cite{inproceedings,Gao2019GraphCT} is another category of used graph inspired neural network in computer vision area. Again, no novelty in architectural designs has been identified within this subfield's contributions.

\subsection{Recognition (object detection, categorization)}
Object detection is a key field in artificial intelligence, allowing computer vision to "see" their environment by detecting objects in visual images or videos. This category encompasses object detection\cite{Luo2020}\cite{Khademi2020}\cite{Gidaris2019}\cite{Kim2019}\cite{MosellaMontoro2019}\cite{Liu2020},image classification\cite{Campos2021},image classification\cite{Marino2017},visual recognition \cite{Li2018BeyondGL}\cite{DBLP:journals/corr/abs-1905-11634} and recognizing ingredients\cite{Chen2020}. New architectures studied in this context were covered below. 

\subsubsection{Cascade Graph Neural Networks}
Cascade Graph Neural Networks (Cas-GNN) processes the two data sources individually and employs a novel Cascade Graph Reasoning (CGR) module to learn powerful dense feature embeddings, from which the saliency map can be easily inferred. Modeling and reasoning of high-level relations between complementary data sources explicitly allow us to overcome challenges such as occlusions and ambiguities better. 

\subsubsection{Deep Generative Probabilistic Graph Neural Networks}
Deep Generative Probabilistic Graph Neural Networks (DG-PGNN) are used inter alia to generate a scene graph for an image. The input to DG-PGNN is an image with a set of region-grounded captions and object bounding-box proposals. Scene graph generation with DG-PGNN constructs and updates a new model called a Probabilistic Graph Network (PGN). A Probabilistic Graph Network can be thought of as a scene graph with uncertainty. It represents each node and each edge by a CNN feature vector and defines a probability mass function for the node-type of each node and edge-type of each edge. The DG-PGNN sequentially adds a new node to the current PGN by learning the optimal ordering in a Deep Q-learning framework, where states are partial PGNs, actions choose a new node, and rewards are defined based on the ground truth. After adding a node, DG-PGNN uses message passing to update the feature vectors of the current PGN by leveraging contextual relationship information, object co-occurrences, and language priors from captions. The updated features are then used to fine-tune the Probabilistic Matrix Factorization (PMFs)\cite{Khademi2020}.

\subsubsection{Edge-labeling graph neural networks}
Edge-labeling graph neural networks (EGNNs) adapt deep learning networks for tasks with few-shot learning \cite{Kim2019}. Neural network-inspired graphs aim to predict the edge labels rather than the node labels on the graphs. Parameters in this network are learned by episodic training with an edge-labeling loss. EGGNs learn to predict edges using intra-cluster and inter-cluster similarities that enable explicit clustering evolution. Learning in EGNNs allows obtaining the generalized model for low-data multiclass problems without retraining.

\subsubsection{Graph Neural Network Denoising Autoencoders}
Denoising Autoencoder network (DAE) takes as input a set of classification weights corrupted with Gaussian noise and learns to reconstruct the target-discriminative classification weights\cite{Gidaris2019}. In this case, the injected noise on the classification weights serves to regularize the weight generating meta-model. DAE model is used as a Graph Neural Network (GNN) to capture the co-dependencies between different classes. 

\subsubsection{LatentGNN}
LatentGNN aims to efficiently encode the long-range dependencies instead of relying on a fully-connected graph defined on input features. LatentGNN was introduced as a solution to Learning Efficient Non-local Relations for Visual Recognition. To construct LatentGNN Restricted Boltzmann Machine (RBM) \cite{RBM} with undirected graph models with latent representations are being used. In LatentGNN, the original feature nodes are augmented with a set of additional latent nodes, and then these nodes are connected structurally.

\subsubsection{Multi-‐Relational Graph Convolutional Network}
Multi-Relational Graph Convolutional Networks (mRGCN) were proposed as a solution to integrate ingredient hierarchy attribute for zero-shot ingredient recognition\cite{Chen2020}. Recognition with mRGCN can develop models that can identify unseen ingredients, as obtaining a fully annotated training set is sometimes too complex due to the high visual variance of input data. mRGCN comprises two modules: mRCGN works for the unseen ingredient prediction, and Deep Convolutional Neural Networks Classifier deals with already known ingredients. Then the obtained trained classifier is used to generate ground-truth supervising the learning process of mRGCN.

\subsubsection{OD-GCN: Object Detection Boosted GCN}
OD-GCN (Object Detection with Graph Convolutional Networks) was proposed to boost the detection performance. The architecture of OD-GCN combines two models: Object Detection (OD) and Graph Convolutional Network (GCN). OD model represents any classical object detection model, so in OD-GCN we can use, e.g., Fast-RNN or others. The second part of OD-GCN is a model based on knowledge graphs that can be used for post-processing.  

\subsubsection{Residual Attention Graph Convolutional Networks}
Residual Attention Graph Convolutional Networks (RAGCNs) constitute graph-based variants of residual neural networks using attention mechanisms and residual solutions to exploit the intrinsic geometric context inside a 3D space. RAGCNs \cite{MosellaMontoro2019} allow the use of organized or unorganized 3D data without using any point features, which increases the effectiveness of the proposed architecture.

\subsection{Scene Analysis and Understanding}
Scene graph is a data structure, which is used for scene generation \cite{Ashual2019SpecifyingOA, 8954048, 10.1007/978-3-030-20876-9_42, 9022385}. Sometimes we want to know more about objects relationships on a scene \cite{9149910}. Graphs neural networks seem to be a natural choice for generating such scene graphs. We have identified one architecture advance within this subfield. 

\subsubsection{Dynamic Gated Graph Neural Network}
Dynamic Gated Graph Neural Network (D-GGNN)\cite{10.1007/978-3-030-20876-9_42} is a deep generative architecture that can be applied to an input image when only partial relationship information, or none at all, is known. It is an upgrade to the Gated Graph Neural Networks (GGNN)\cite{2015Yujiaarxiv:1511.05493}. This latter network type cannot work when input data is partial, which is a severe limitation in computer vision where missing data often occurs. D-GGNN, similarly to Generative Probabilistic Graph Neural Networks (DG-PGNN) are frequently used for scene graph generation tasks. This architecture builds the graph in the process of reinforcement learning (RL). The graph built so far in each training step is the current state. The D-GGNN leverages the power of GGNN to exploit the relational context information of the input image and combines the gated graph neural network with an attention mechanism.  

\subsection{Segmentation, grouping, and shape}
In this area, we keep focusing on usability and the possibility to apply Graph Neural Networks in segmentation, grouping, and shape area. In this subfield, we have identified one novel GNN structure aimed at addressing the problem of referring expression comprehension in videos \cite{DBLP:journals/corr/abs-1812-04794}.

\subsubsection{Language-Guided Graph Attention Networks}
Language-Guided Graph Attention Networks (LGRANs) highlight the relevant content referred to by the expression. The graph attention comprises two main components: a node attention component to highlight relevant objects and an edge attention component to identify the object relationships present in the expression. In LGRANs additionally, encode the relationships between objects (that have properties of their own). Secondly, in LGRANs, attention is guided by the referring statement. Thirdly, LGRANs, unlike GATs, maintain different types of features to represent the node properties and node relationships.

\subsection{Video analysis and understanding}
In this subfield we have identified the application of GNN in Video understanding models\cite{10.1007/978-3-030-01228-1_25} or video anomaly detection under weak labels\cite{DBLP:journals/corr/abs-1903-07256}. Three specific architectural types are covered below.

\subsubsection{Curve-GCN}
Curve-GCN is used for interactive object annotation, using parametrized objects with either polygons or splines, and is trained end-to-end at a high output resolution. Curve-GCN alleviates the sequential nature of PolygonRNN, by predicting all vertices simultaneously using a Graph Convolutional Network (GCN)\cite{DBLP:journals/corr/abs-1903-06874}.

\subsubsection{Social Relationship Graph Generation Networks}
Social Relationship Graph Generation Network (SRG-GN) proposed by Goel et al\cite{Goel2019} uses memory cells like Gated Recurrent Units (GRUs) to iteratively update the social relationship states in a graph using scene and attribute context. SRG-GN can unify the representation of social relationships and attributes forming input images. This architecture is based on recurrent connections among the GRUs. They were implemented for passing messages between nodes and edges, improving GRUs social relationship recognition.

\subsection{Vision applications and systems, vision for robotics and autonomous vehicles}
It is one of the two most extensive areas of computer vision. Research in this category is more application-oriented and focuses on solutions for available problems rather than architectural advances and preparing complex solutions. In this category, we can find research studies dealing with modeling edge features with Deep Bayesian Graph Networks\cite{Atzeni2021}, Graph Convolutional Gaussian Processes \cite{DBLP:journals/corr/abs-1905-05739}, GCAN Graph Convolutional Adversarial Network \cite{Ma2019}, data representation and learning with Graph Diffusion-Embedding Networks \cite{Jiang2019}, extracting building polygons\cite{Wei2022} or temporal prediction\cite{Kostrikov2018}. Three architectural advances in this subfield will be named below.

\subsubsection{Graph Convolutional Adversarial Networks}
Graph Convolutional Adversarial Networks (GCANs) were proposed for unsupervised domain adaptation. GCAN \cite{Ma2019} is a model which uses three kinds of inputs: structure, domain label, and class label to generate the deep model for unsupervised domain adaptation. GCANs employ structure-aware alignment, domain alignment, and class centroid alignment to reduce the domain discrepancy for domain adaptation. 

\subsubsection{Graph Diffusion-Embedding Networks}
Graph Diffusion-Embedding Networks (GDENs) were proposed as a method of graph-based feature diffusion. GDENs\cite{Jiang2019} integrate both feature diffusion and graph node embedding simultaneously into a unified network. By employing a novel naturally scalable diffusion-embedding architecture with diffusion-embedding operation, a simple closed-form solution can be obtained, which thus guarantees high efficiency.

\subsubsection{Graph Convolutional Gaussian Processes} 
Graph Convolutional Gaussian Processes \cite{DBLP:journals/corr/abs-1905-05739} represent a novel Bayesian nonparametric method in non-Euclidean domains. Graph Convolutional Gaussian Processes can be applied to problems in machine learning, for which the input observations are functions with domains on general graphs.

\subsection{Vision + other modalities, Vision + language}
Second largest of category of CVPR in terms of number of contributions is a Vision + other modalities, Vision + language. In this category we have identified papers on image annotation\cite{Zhang2019, Kumar2018, Goel2019, Acuna2018}, visual question answering \cite{2015Jacobarxiv:1511.02799, 2016Damienarxiv:1609.05600, 2018Willarxiv:1806.07243, 2018Medhiniarxiv:1811.00538, 2020Dengarxiv:2008.09105, Jiang2020}, situation recognition\cite{Suhail2019, Li2017}, as well as image understanding, image captioning and object annotation\cite{DBLP:journals/corr/abs-1903-06874}. In this subfield no novelty in architectural designs have been identified within GNN-based contributions.

\subsection{Mesh Generation and Reconstruction}
This category was introduced to accommodate studies in mesh generation \cite{Daroya2020,Dai2019, Tang2019, Wang2018} and reconstruction\cite{DBLP:journals/corr/abs-1901-11461} in which graph inspired neural networks can be naturally employed.

\section{Datasets} \label{sec:datasets}
Contributions presented in this paper cover a variety of applications which require for their validation benchmarks of different nature. Still, there exist a common set of benchmarks which are omnipresent in the GNN-based computer vision studies. The list of most used datasets, along with their field of appliactions and characteristics, is presented in Table \ref{tab:databases}.

\begin{table}
    \centering
    \caption{Most used datasets in  the GNN-based computer vision studies.}
    \resizebox{\columnwidth}{!}{
    \begin{tabular}{c|c|c} 
       Dataset name         & Fields of application & Details              \\ \hline
       NTU-RGBD             & Action and behavior recognition  & 60 action classes and 56,880 video samples            \\ \hline
       CUHK03               & Biometrics, face, gesture, body pose  & 14,097 images         \\ \hline
	Skeleton-Kinetics    & Action and behavior recognition  &     300,000 videos\\ \hline
       Visual Genome        & Scene analysis     & 108,077 images \\ \hline
       Visual Genome        & Low-level and physics-based vision     & 108,077 images \\ \hline
       Visual Genome        & Mesh Generation and reconstruction     & 108,077 images \\ \hline
       ShapeNet             & Mesh Generation and reconstruction  & 55 common object categories   \\ \hline
       ShanghaiTech         & Video analysis and understanding    & 1198 annotated crowd images \\ \hline
      COCO                  & Recognition (object detection, categorization)  & over 320,000 images   \\ \hline
       CoNSeP & Medical, biological and cell microscopy & 41 HE stained image tiles    \\ \hline
       Colorectal Cancer (CRC)    & Medical, biological and cell microscopy   & 139 images \\ \hline
       MS-Caleb-1M          & Computational photography, image and video synthesis & 10 million face images   \\ \hline
	Jilin-1 and SkySat       & Motion and tracking   & 8 satellite video sequences  \\ \hline
    \end{tabular}
    }
    \label{tab:databases}
\end{table}

\section{Relationships and Inspirations}\label{sec:inspiration}
\begin{figure}[htbp]
    \centering
    \includegraphics[width=\columnwidth]{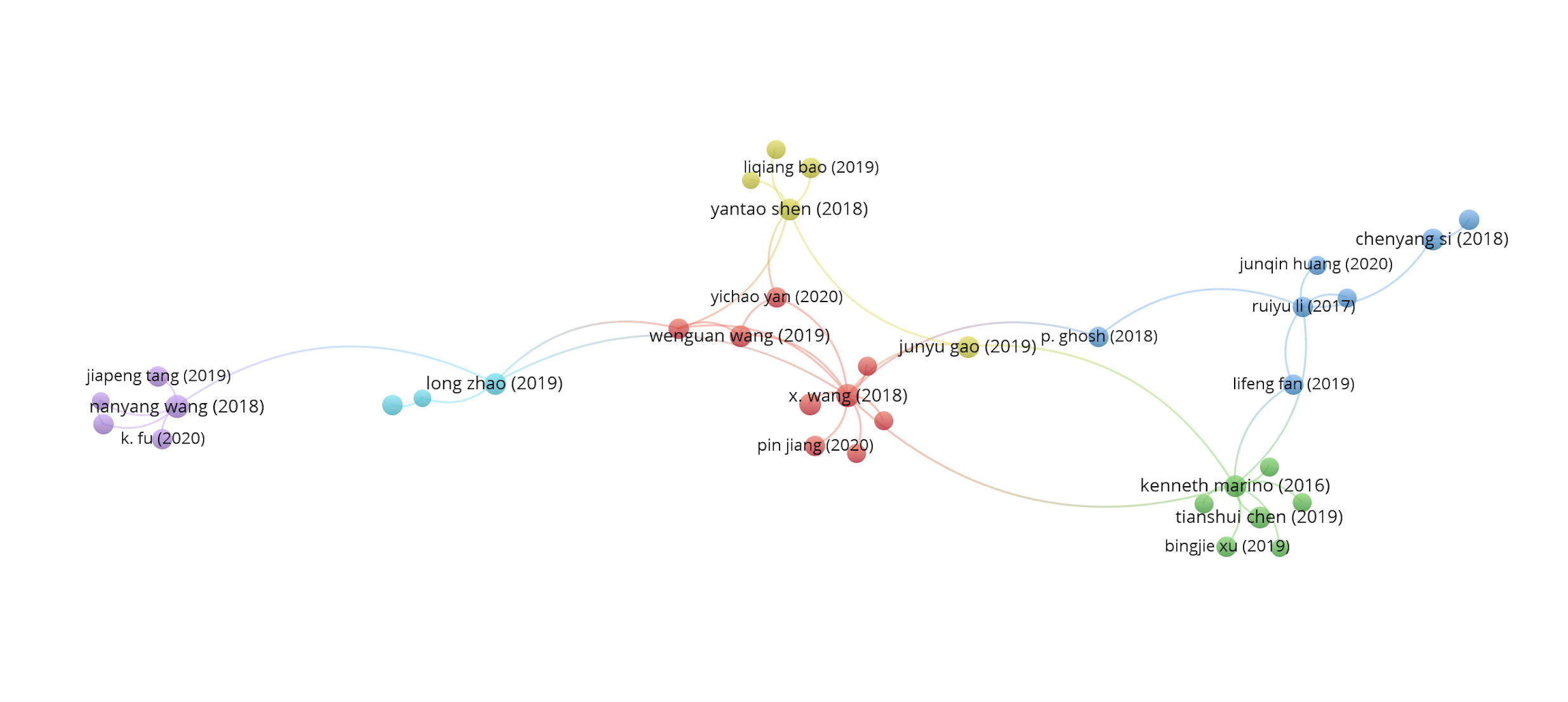}
    \caption{Visualization of links between GNN-based computer vision papers}
    \label{fig:graph1}
\end{figure}

To visualize relations between manuscripts described in this paper, we used a graph-based approach. It was employed to show relations between papers we have covered in our study. Relationships are represented by edges and publications by nodes. Marker size corresponds to the number of times the paper has been cited in the other articles from the analyzed set. Connections between articles and references are based on search engine and API of \href{https://www.semanticscholar.org}{Semantic Scholar} which indexes publications published by major scientific publishers (including IEEE, ACM, Springer, or Wiley). The resulting graph is presented on Figure \ref{fig:graph1}. 

The resulting graph is presented on Figure \ref{fig:graph1}. It can be seen that clusters of articles, similar to the groups of CVPR conference topics, can be identified. It allows us to distinguish papers from similar methodological backgrounds, though often not sharing the same scope or architecture. We can also identify papers present in the graph (e.g. \cite{Marino2017,Wang2018}) which have significant impact on publications from another clusters. This also concerns the major publications in another clusters (like \cite{10.1007/978-3-030-01228-1_25} has impact on \cite{Wang2018}).

\begin{figure}[htbp]
    \centering
    \includegraphics[width=\columnwidth]{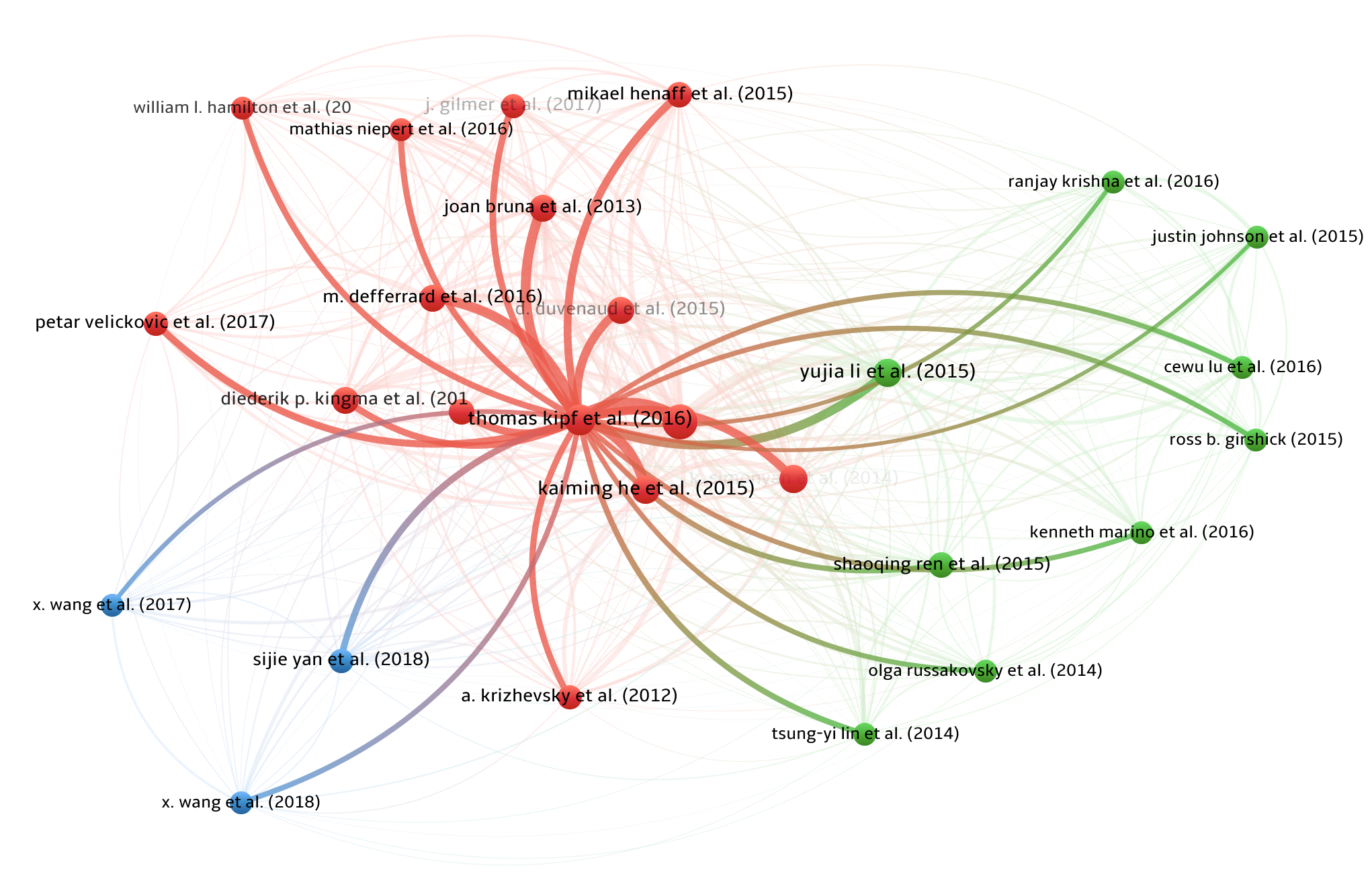}
    \caption{Visualization of the impact of the Kipf et al paper "Semi-Supervised Classification with Graph Convolutional Networks in computer vision"}
    \label{fig:graph2}
\end{figure}

\begin{figure}[htbp]
    \centering
    \includegraphics[width=\columnwidth]{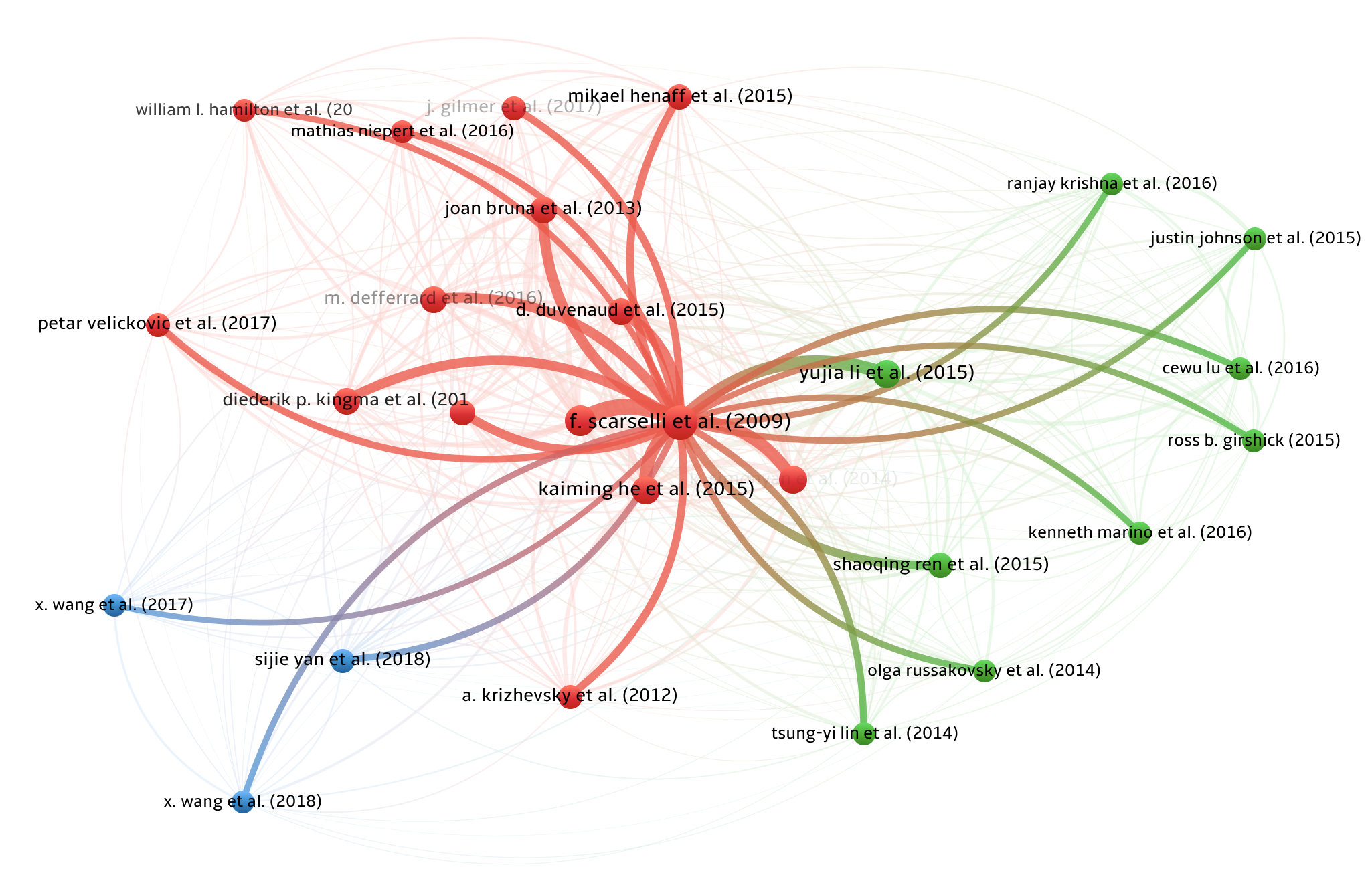}
    \caption{Visualization of the impact of the Scarselli et al paper "The Graph Neural Network Model in computer vision"}
    \label{fig:graph3}
\end{figure}

On Figures \ref{fig:graph2} and on \ref{fig:graph3}  we have demonstrated how significant impact for the analyzed publications -- measured by citation numbers -- have other articles, not only in the domain of computer vision. The larger the size of the marker of a given article, the more significant the impact. We can see how big impact have on computer vision based on graph neural network have fundamental contributions \cite{4700287,Kipf2017SemiSupervisedCW} prepared by F.Scarselli et al and by Kipf et al. Both graphs' source codes are made available in the  \href{https://github.com/mkrzywda/Graph-Neural-Networks-in-Computer-Vision---Architectures-Datasets-and-Common-Approaches}{repository} to allow their replication. 

\begin{table*}[htbp]
    \centering
    \caption{Highly-influential scientific articles for the development of GNN-based computer vision studies}
    \begin{tabular}{c|c|c}
        Title of articles & References count & Reference \\ \hline
            The Graph Neural Network Model & 167 & \cite{4700287} \\ \hline
            Semi-Supervised Classification with Graph Convolutional Networks & 126 & \cite{Kipf2017SemiSupervisedCW} \\ \hline
            Deep Residual Learning for Image Recognition & 94 & \cite{7780459} \\ \hline
            Gated Graph Sequence Neural Networks & 82 & \cite{2015Yujiaarxiv:1511.05493} \\ \hline
            Very Deep Convolutional Networks for Large-Scale Image Recognition & 65 & \cite{Simonyan2015VeryDC} \\ \hline
            Convolutional Neural Networks on Graphs with Fast Localized Spectral Filtering & 64 & \cite{10.5555/3157382.3157527} \\ \hline
            Faster R-CNN: Towards Real-Time Object Detection with Region Proposal Networks & 63 & \cite{Ren2015FasterRT} \\ \hline
            Adam: A Method for Stochastic Optimization & 63 & \cite{Kingma2015AdamAM} \\ \hline
            Convolutional Networks on Graphs for Learning Molecular Fingerprints & 58 & \cite{Duvenaud2015ConvolutionalNO} \\ \hline
            Spectral Networks and Locally Connected Networks on Graphs & 55 & \cite{Bruna2014SpectralNA} \\ \hline
            A new model for learning in graph domains & 50 & \cite{Gori2005ANM} \\ \hline
            Microsoft COCO: Common Objects in Context & 49 & \cite{Lin2014MicrosoftCC} \\ \hline
            Graph Attention Networks & 45 & \cite{vel2018graph} \\ \hline
            Neural Message Passing for Quantum Chemistry & 43 & \cite{Gilmer2017NeuralMP} \\ \hline
            ImageNet Large Scale Visual Recognition Challenge & 39 & \cite{Russakovsky2015ImageNetLS} \\ \hline
            Deep Convolutional Networks on Graph-Structured Data & 37 & \cite{Henaff2015DeepCN} \\ \hline
            Inductive Representation Learning on Large Graphs & 37 & \cite{Hamilton2017InductiveRL} \\ \hline
            Non-local Neural Networks & 36 & \cite{Wang2018NonlocalNN} \\ \hline
            Fast R-CNN & 35 & \cite{Girshick2015FastR} \\ \hline
            Mask R-CNN & 34 & \cite{He2020MaskR} \\ \hline
    \end{tabular}
    \label{tab:articles}
\end{table*}

In Table \ref{tab:articles} we have presented the top 20 most cited works within studied articles. After analysis of these references, we can see that in addition to the base article in the field of graph neural networks \cite{4700287}, issues related to, inter alia, with recursive networks \cite{He2020MaskR} \cite{Girshick2015FastR} \cite{Ren2015FasterRT} had a significant impact on the development of Graph Neural Networks in the area of computer vision. Moreover, we can see that the graph's attention mechanism \cite{vel2018graph} and convolutional operations on graphs \cite{Kipf2017SemiSupervisedCW} had a significant influence on the development of Graph Neural Networks-based methods in this field. 

\section{Conclusion}
In the last few years, studies on GNNs have become an active research field, mainly due to the advance of underlying Deep Learning architectures. The essential idea of GNNs is to enhance the node representations by propagating information between nodes. This paper provided an analysis of state-of-art Graph Neural Networks in computer vision. We delivered a thorough review of the latest achievements, summarizing architectural advances. We also studied the relations between methods and identified sources of inspiration and related studies, also outside of the explored domain. We have discovered that studies on skeleton-based action recognition and research in one subfield  -- biometrics, face, gesture, body pose -- seem to be very influential for developing vision systems using GNNs. We anticipate that the results of this study might be useful for researchers, not only in the field of computer vision.

\section*{Acknowledgments}
The work was supported by the Faculty of Physics and Applied Computer Science AGH UST statutory tasks within the subsidy of MEiN.

\bibliography{Krzywda_et_al_WCCI2022}
\bibliographystyle{IEEEtran}

\end{document}